# CT-image Super Resolution Using 3D Convolutional Neural Network

**Yukai Wang, Qizhi Teng,[*] Xiaohai He, Junxi Feng, and Tingrong Zhang**

Sichuan University, College of Electronics and Information Engineering, No. 24 South Section 1, Yihuan Road, Chengdu 610065, China

**Abstract.** Computed Tomography (CT) imaging technique is widely used in geological exploration, medical diagnosis and other fields. In practice, however, the resolution of CT image is usually limited by scanning devices and great expense. Super resolution (SR) methods based on deep learning have achieved surprising performance in two-dimensional (2D) images. Unfortunately, there are few effective SR algorithms for three-dimensional (3D) images. In this paper, we proposed a novel network named as three-dimensional super resolution convolutional neural network (3DSRCNN) to realize voxel super resolution for CT images. To solve the practical problems in training process such as slow convergence of network training, insufficient memory, etc., we utilized adjustable learning rate, residual-learning, gradient clipping, momentum stochastic gradient descent (SGD) strategies to optimize training procedure. In addition, we have explored the empirical guidelines to set appropriate number of layers of network and how to use residual learning strategy. Additionally, previous learning-based algorithms need to separately train for different scale factors for reconstruction, yet our single model can complete the multi-scale SR. At last, our method has better performance in terms of PSNR, SSIM and efficiency compared with conventional methods.

*Qizhi Teng, E-mail:qzteng@scu.edu.cn

## 1 Introduction

CT is a three-dimensional (3D) imaging technique which is widely used to provide detailed information for accurate analysis. Recently, CT, micro-CT and nano-CT have been the popular equipment to display real 3D rock sample images.[1] Establishment of accurate 3D image of rock can provide rich structure information to help geological researchers analyze the physical properties of rocks[2,3] and play an important role in the field of geological and petroleum exploration. As shown in Fig. 1, a complete 3D-CT image is actually composed of two-dimensional (2D) slice images. Due to its inherent limitations of CT devices, setting high resolution will not only need high cost, but will result in decrease of field of view (FOV), causing the loss of long-range properties of reservoirs rock.[11] In many cases, there are only LR CT images available for analysis. Therefore, the use of super resolution(SR) algorithm is an effective method to improve the resolution of CT images, which can provide more clear sample data for subsequent geological research or medical diagnosis.

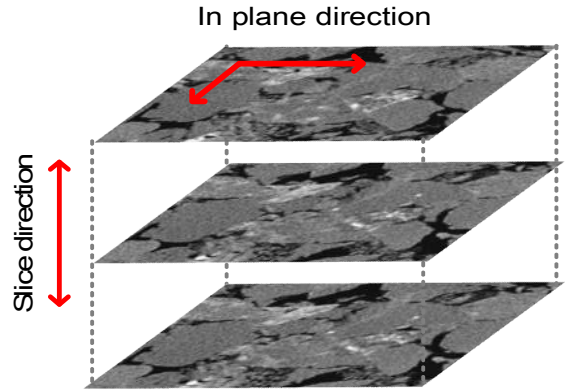

**Fig. 1** CT images acquisition. 3D image acquisition is inconvenient, usually composed of a series of 2D slice images.

SR reconstruction, having drawn extensive attention in computer vision field, is a handy method to improve quality of image.[6] If external examples are given, learning-based SR algorithms are more plausible to acquire good results. Deep learning techniques[12] recently have shown remarkable performance in the tasks of image classification[16], object detection[17], etc., and is superior to conventional machine learning algorithms. Dong et al. introduced deep learning to SR[4] and raised a network SRCNN[5] which only contains 3-layers CNN structure but outperforms former methods. Subsequently, Jiwon Kim found that deeper network structure shows a significant promotion and proposed VDSR[13] to resolve issues in Dong's work.

In current research, scholars mainly focus on single 2D image SR rather than spatial 3D voxel. Specially, SR researches for 3D images are aiming at restoring magnetic resonance imaging (MRI). H. Greenspan et al. achieved SR of MRI images in slice direction using an iterative algorithm.[8] Manjon proposed a non-local MRI upsampling method[7] where some of underlying high frequency information can be recovered. Iwamoto proposed a method based on sparse representation and self-similarity to improve resolution of MRI,[9] which only improves the resolution in the slice direction, and have no effect in plane direction.

Yuzhu Wang[10] used neighbor embedding algorithm to improve resolution of CT image of rock samples, in the meanwhile, high frequency information was supplemented by high resolution scanning electron microscopy (SEM) image. Li proposed a voxel SR reconstruction algorithm[11] based on sparse representation, which can improve the resolution in all directions.

Zhang et al. extended adjusted anchored neighborhood regression algorithm (A+)[14], to 3D and proposed high frequency modified 3DA+ algorithm[15], where a correlative dictionary and mapping matrix between high frequency and low frequency was established. In reconstruction stage, the matched dictionary atom and mapping matrix were searched for each input of the 3D block to complete SR.

Unfortunately, the aforementioned algorithms are focused on 2D images, in view of the fact of 3D-CT images of rock, the following issues remain to be solved: First, the computational intensity and memory of 3D image data is far greater than the 2D images, so the method to handle with 2D images can't be directly transferred to 3D model; Second, CT samples are not as convenient as 2D images to obtain, that is to say, it's not easy to get substantial alignments of rock CT samples to training network. In addition, CT image of rock has the characteristics of low contrast, single texture, and complex pore structure, which all bring difficulty to task of SR; Third, during training network and reconstruction stage, the calculation and time complexity have to be taken account to ensure our work can be carried out on the general computing equipment. Hence, it is desirable to devise a new network to cope with SR for voxel images.

In order to enhance resolution of CT images of rock from three directions (i.e., x, y ,z), we propose a novel network, termed as 3D super-resolution convolutional neural network (3DSRCNN), to promote resolution for volumetric images. Before training, cropping initial samples of large size to sub-blocks is necessary for a promising result and we give reasonable explanation about it. Deeper network architecture may cause gradient exploding and slow convergence, so we employed some feasible strategies, including residual learning, gradient clipping, and adjustable learning rate, etc., to optimize training process. Experiments show the proposed network can be applied to different scale factors and performs equally to method that separately train network with different scale.

In summary, we introduce 3DSRCNN to realize the SR reconstruction of 3D CT images. Furthermore, we have experimentally investigated the influence of network depth on the reconstruction quality. Thus employing a moderate number of network layers is of significance. Subsequently, we demonstrate that it is necessary to use the residual learning when the number of network layers goes deeper. Consequently, we make corresponding adjustments on network architecture and training strategies, so as to achieve a trade-off between the accuracy and speed. Moreover, we have addressed the problems existing in pre-processing CT images such as consuming too much memory, etc. The proposed 3DSRCNN have achieved the state-of-art performance in terms of PSNR and SSIM, while ours have fairly faster reconstruction speed on GPU.

The remainder of paper is organized as follows. In Sec. 2, we firstly introduced the concept of SR and the implementation of deep learning on it. Then, in Sec. 3, the proposed network--3DSRCNN was described in detail. In Sec. 4, we investigated how to design the network and pursuit better performance of accuracy and speed by experiments. Besides, we also tested and compared 3DSRCNN with state-of-the-art methods. In Sec. 5, conclusion and future studies of our work were given.

## 2　Related Works

In this section, the conception of SR and the corresponding method of using CNN will be briefly introduced.

### 2.1　Image Super-Resolution

Single image super resolution (SISR) is an ill-posed problem due to lacking of detailed information. There are two traditional methods to recover low resolution (LR) images to high resolution (HR) images, one is using context correlation in LR image yet has inborn defects that it cannot acquire more specific high frequency information; The second is the learning-based method that can acquire the prior information via training given images. The process of SISR is interpreted as following, for a ground truth image (i.e. HR images) $X$, and LR images $Y$.

$$X \approx F(Y) \qquad\qquad (1)$$

Here, we try to find a function as $F(Y)$, which can restore LR to HR to a certain extent. It is an under-determined problem and most of recent state-of-art methods adopted learning based strategy to solve it. SR reconstruction based on learning method is to learn the mapping relation between low frequency information and high frequency information by iteratively training paired LR and HR images. The classical ones such as sparse-representation method[17,19], is basically composed of three steps: (1) LR features extraction; (2) Learning the mapping relation between LR and HR patch; (3) Reconstruction of HR images using learned mapping relation. In A+ algorithm[14], the non-linear mapping relationship from low resolution space to high resolution space is transformed into mapping matrix instead of continuous iteration for optimal solution, which can be realized by deep learning technique.

## 2.2    *Convolutional Network for Super-Resolution*

Dong et al. considered that deep convolutional neural network is equivalent to the aforementioned pipeline, which can directly learns an end-to-end mapping relation. While SRCNN have achieved good result in 2D image datasets, there are still limitations as following: First, its single model works only for single scale, which cannot be applied on different upscaling factors; Second, training of SRCNN converges too slowly.

Jiwon Kim introduce VDSR that adopt a deeper network structure of 20 layers to break through the limitations in SRCNN and pointed out stacking more CNN layer allows convolutional filters to become increasingly global, which conceptually benefits to learn mapping relation. In Ref. 13, they have experimentally validated the viewpoint –'the deeper, the better'. The SR technique of single 2D image has been very mature, but these can't be directly converted into a 3D model. Because the amount of data used to calculate in 3D image is far larger than the 2D image, it's necessary to redesign network architecture. Furthermore, acquisition of 3D image sample is not as easy as 2D images. We try to use a relatively small samples to complete the training of the network as far as possible.

## 3    Three-dimensional Super-Resolution Convolutional Neural Network

In this Section, we introduce the structure of the 3DSRCNN that consist of 12-layers of 3D-CNN. Besides, some strategies for optimizing training process are employed to our network 3DSRCNN. Next, we describe production of training data in detail.

### 3.1    Network Structure

We proposed the 3D network structure, named as 3DSRCNN, to achieve SR for volumetric CT images as shown in Fig. 2.

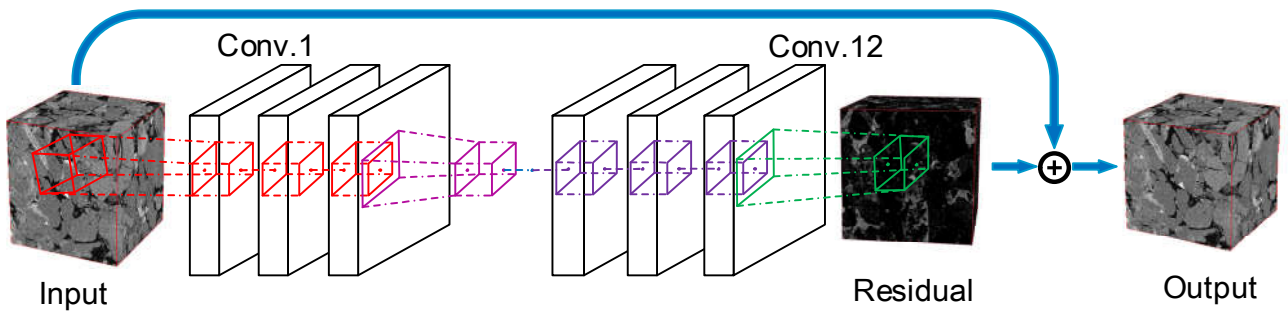

**Fig. 2** Network structure of 3DSRCNN contains 12 layers 3D-CNN which exploit 3D spatial information. Each 3D-CNN has 64 filters to capture diverse features. LR image goes through layers and transforms into an HR image. Whole 12-layers of network actually yields the prediction of the residual image. We combine residual image and input as the final output as HR image.

For volumetric super resolution, we employ a network composed of 12 layers each of that has 64 channel filters (i.e. convolutional kernel). The first layer is responsible to extract low frequency patch from LR images; The middle 10 layers learns mapping relationship between LR and HR volumetric block; The last layer combine learned mapping relation and initial LR images to finally formulate SR images.

The convolution network actually extracts spatial correlative information which contains diverse pattern features. Recent study[20,22,23] shows increasing depth using an architecture with very small convolution filters (3×3), which shows that a significant improvement on image recognition, etc. Simultaneously, when the input image continues to pass through the CNN, the extracted feature becomes global and has a larger receptive field. Consequently, the depth of network layers will affect the reconstruction accuracy and training time. Due to original images containing rich texture information, the deeper network structure has better SR ability as Jiwon pointed in their work[13]. Computation complexity, however, is a non-negligible topic which directly influence the practical application of our algorithm. The whole computation complexity of network can be calculated as

$$\sum_{l=1}^{D} M^3 \bullet K^3 \bullet C_{l-1} \bullet C_l \tag{2}$$

Where $D$ is the depth of CNN layers, $l$ identify the current layer number, $C$ is the number of channels, $M$ is the feature map size. It is obvious to find that dense network structure would increase the computational complexity.

SRCNN has no padding before convolutional operation, causing boundary pixel missed. On the contrary, padding is necessary for our network because that processing 3D image will typically occupy a lot of memory. In order to save memory, we divide the initial CT image to sub-blocks with small size. Given that the layers of network is 12, the input size is relatively small, which will cause majority loss of internal information without padding during forward propagation. Hence, we use zero padding and subsequent experimental have proved the correctness of this scheme.

Because the SRCNN network has only three layers of network, it not only completes learning mapping relation between LR and HR but also remains initial LR feature during forward propagation. When largely increase layers, the information of the input LR feature will be lost in the continuous convolution process, which leads to training unstable and discard initial information. We consider residual learning can be used to solve the above problems. After each CNN, we utilize Rectified Linear Unit (ReLU)[30] as activation function on output of last layer.

$$\mathrm{Re}\, LU(X) = \max(0, WX + b) \tag{3}$$

Where $X, W$ denote input and weight parameter respectively, $b$ is bias.

## 3.2    Pre-process of Training-Set

Before training, we should crop and transform initial CT sample to suitable shape. Specifically, the step of crop original CT samples is introduced as follows.

We separately use factor $=2,3,4$ to downsample ground truth CT datasets $\{Y\}$ in different samples, then we use bicubic interpolation to upsample them by same size and these are used as LR images $\{X\}$. Original CT images are cropped to sub-3D-blocks for producing training set. HR images $\{Y\}$ is viewed as label to calculate loss function, and LR images $\{X\}$ are fed in network. The whole process is shown in Fig. 3.

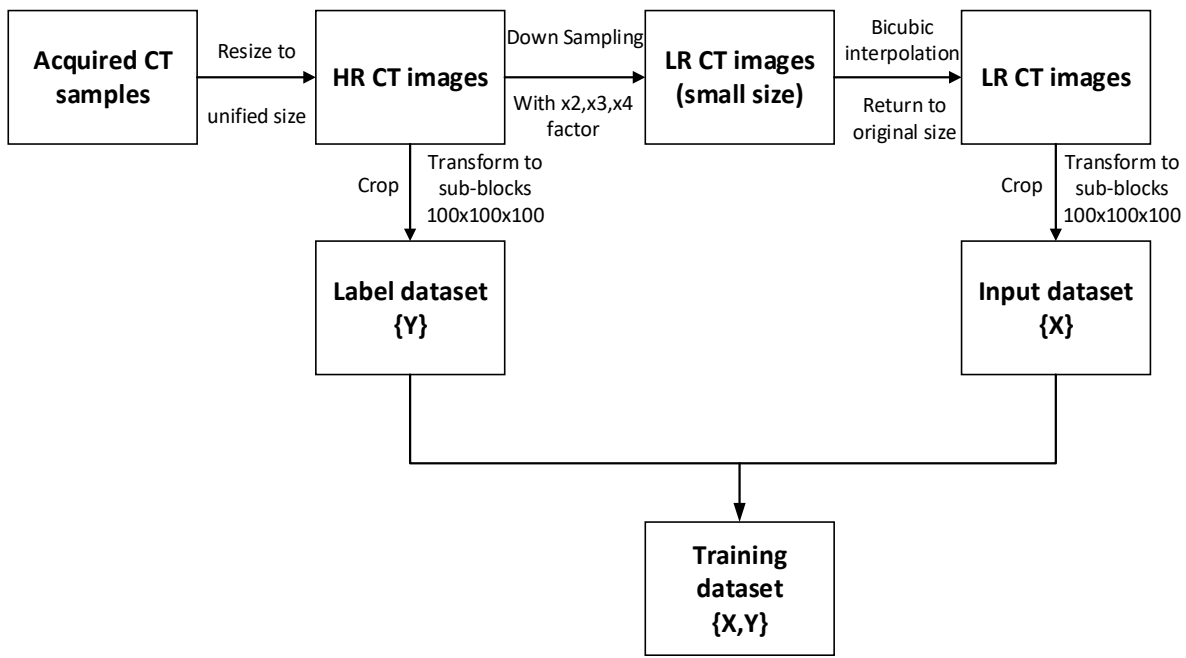

**Fig. 3** Procedure of producing training set.

When deal with CT images, crop is significant for training and there are mainly three points as following:

(1) In this way, a larger quantity of training samples can be obtained via image cropping under the condition of limited number of CT samples. These sub-blocks are viewed as small size 'images' rather than 'patch'.

(2) Cropping promises our program running in general computer since training 3D-block will occupy amounts of memory. When large CT block is cropped into small blocks, it enables computing devices to calculate under low load.

(3) The sub-blocks are overlapping containing redundant information, in the sense that training set have rich contexture that is advantageous to learn mapping relationship.

Assuming that the input is a cubic block, the specific number of training samples after cropping can be counted with following: $(\frac{I_{inp}}{stride + I_{sub}})^3$, Where $I_{inp}$ is the length of a single side of initial images, $I_{sub}$ is sub-block's length, $stride$ is span length when cropping. Cropping initial images to small $I_{sub}$ can produce a large datasets, but it will also cause LR feature pattern incomplete.

After the cropping, datasets is composed of paired $\{X, Y\}$ that are used as input and label for training, respectively. Experiments show that setting $I_{sub}$ as 25 is appropriate.

### 3.3 Training Strategies

Our proposed network is constituted of massive tensors which represent end-to-end mapping relationship. Weight parameters $\Theta$ of network is initialized by Gaussian distribution (zero mean and standard deviation 0.001). Through continuous iterative training, $\Theta$ is increasingly optimized by Mean Squared Error (MSE) loss function. However, directly using standard stochastic gradient descent (SGD) takes long time to converge. We employ some strategies to optimize our network structure and training data.

### (1) Residual Learning

In spite of stacking more layers may have significant effects, the vanishing/exploding gradients problem will emerge.[24,25] We find MSE error would suddenly increase in a certain training iteration when depth exceed 10 layers. In addition, deeper layer model produce higher training error which makes training process unstable. On the other hand, In Ref. 13, author consider that input detail is discarded after passing convolutional operation in deep layers, which gives birth to that the output only use learned features to generate images. He Kaiming et al. have introduced a deep residual learning framework[26] to and got excellent scores in task of images recognition. Residual-learning strategy is also adopted in our network to solve these problems. We define input as $x$, output as $y$, and residual image $r = y - x$, where $f(x)$ denotes the output of data passing through network. Given training set $\{X, Y\}$, and loss function based on MSE is interpreted as following:

$$L(\Theta) = \frac{1}{n} \sum_{i=1}^{n} \| r - f(x_i; \Theta) \|^2 \qquad (4)$$

Where $n$ is the number of training batch samples. One point must be stressed is that residual learning is not necessary in all cases. When the number of layers is not deep, the use of the residual network have no obvious effect, or degrade instead.

*(2) Adjustable Learning Rate*

In SRCNN, it is found that the training with small learning rate converges very slowly. High learning rate help boost training yet can lead to gradients exploding. We apply the adjustable learning rate strategy for avoiding gradient exploding yet speeding up training. In early epochs, setting relatively high learning rate will be beneficial for accelerating training process. As training epoch going on, learning rate is reduced with following rules.

$$lr = lr * 0.1^{(\frac{epoch}{step})} \qquad (5)$$

Where $epoch$ counts current training times, and $step$ is predefined to control decay of learning rate.

*(3) Momentum Acceleration*

Due to the magnitude of complexity in dealing with 3D images, the convergence of using standard SGD[28] is very slow. We employ momentum SGD to accelerate training process. Momentum is a commonly used acceleration technique in gradient descent. It accumulates the momentum before it replaces the real gradient. The implementation of SGD with Momentum in our work subtly differs from Sutskerver's work.[29] Considering the specific case of momentum, the gradient update formula is written as a new form:

| Algorithm：Momentum SGD |
|---|
| Require：learning rate $lr$，momentum coefficient $\rho$，weight parameters $\Theta$, velocity $v$；<br>while do:<br>    Batch samples：LR images $\{x_1, x_2, \cdots, x_m\}$，residual images as $\{r_1, r_2, \cdots, r_m\}$；<br><br>    Update Gradient: $g \leftarrow \dfrac{1}{m}\nabla_\Theta \sum\limits_{i=1}^{m} L(x_i, r_i, \Theta)$;<br><br>    Update velocity: $v \leftarrow \rho * v + g$;<br>    Update weight parameters: $\Theta \leftarrow \Theta - lr * v$;<br>end |

Where $L(x)$ denotes MSE loss function, and $\Theta, g, v, \rho$ are the weight parameters in network, gradients, velocity, and momentum factor, respectively. In our experiments, momentum factors are all set to 0.9.

*(4) Gradient Clipping*

Gradient clipping[21] is usually applied in training RNN network in case of gradient exploding/vanishing. One simple way is pre-defining a threshold to clip them whenever they go over the threshold. In VDSR[13], they use this technique to limit gradients to a certain range. In our work, we directly clip gradients to range $[-\theta, \theta]$, where $\theta$ is predefined clipping range.

# 4 Experiments and Results

In this Section, we first introduces experimental basis and evaluation metric, peak signal-to -noise ratio (PSNR) and structural similarity index (SSIM), which are widely used to assess image quality. Next, we investigate the influence of important parameters on the accuracy of reconstruction, and analyze the reasons. At last, extensive experiments are conducted to compare our method with others.

## 4.1 Experimental Datasets and Evaluation Criteria

Deep learning generally benefits from big data training, considering the actual situation, it is not easy to get rock CT images, we attempt to use a relatively small number of CT samples to making training set. In order to make the experimental results more convincing, we selected a batch of training samples which come from diverse rock types with representative characteristics. The displayed in Fig. 4 are training CT images of rock samples. The test samples are different from training set which both are consistent with the identical selection rules to guarantee convinced results as shown in Fig. 5. Each selection of CT images has dimensions of 400×400×400 pixels, and each pixel equals to actual length ( $\mu m$ -level) as demonstrated in Fig. 4.

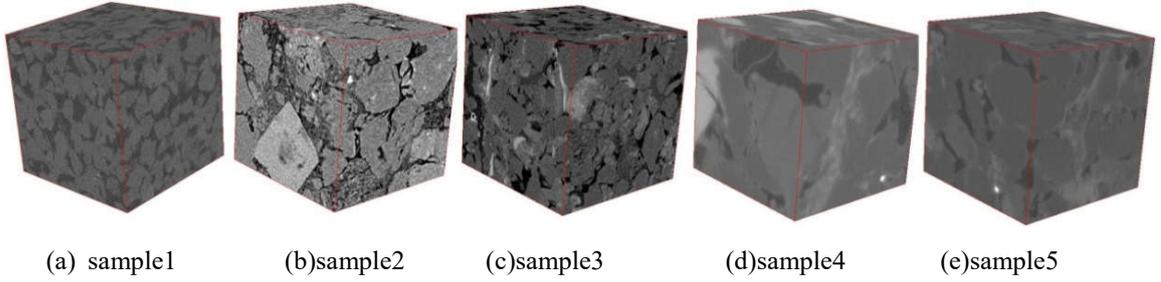

(a) sample1    (b)sample2    (c)sample3    (d)sample4    (e)sample5

**Fig. 4** Five sets of original CT images of rock as training CT samples: (a)(b)(c)are Sandstone with resolution of 3.8 $\mu m$
(d) is carbonate rock with resolution of 1.07 $\mu m$  (e) is sandstone with resolution of 1.07 $\mu m$

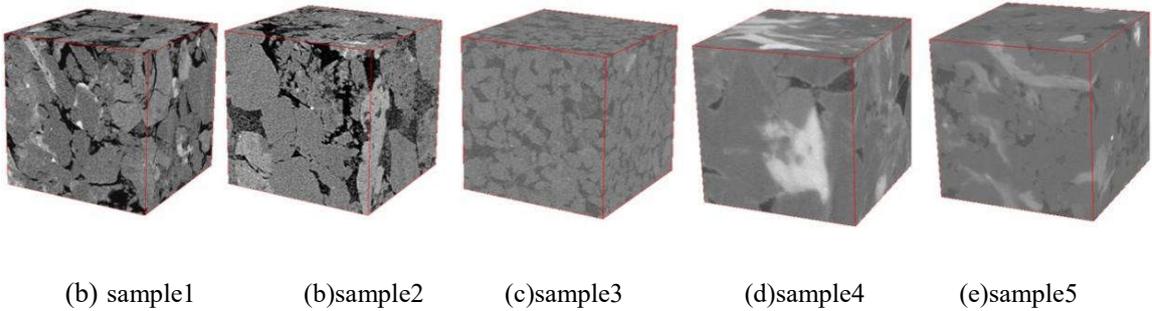

(b) sample1    (b)sample2    (c)sample3    (d)sample4    (e)sample5

**Fig. 5** Five sets of original CT images of rock as testing CT samples.

Results of reconstruction are validated by PSNR and SSIM. PSNR is widely used to measure the quality of image restoration. For a 3D image $f(x, y, z; \Theta)$ with a size of $D \times H \times W$ , PSNR is calculated as follows:

$$MSE = \sum_{z=1}^{D} \sum_{y=1}^{H} \sum_{x=1}^{W} \frac{[\hat{f}(x, y, z) - f(x, y, z; \Theta)]^2}{D \times H \times W} \quad (6)$$

$$\text{PSNR} = 10 \log_{10}(\frac{MAX_f^2}{MSE}) \quad (7)$$

Where $\hat{f}(x, y, z)$ denotes original HR images, $MAX_L$ is maximum gray level. The higher PSNR value indicates that reconstructed CT images are more similar with ground truth. PSNR is based on pixel error, however, it does not take into account the visual features of human eyes. Wang et al. use SSIM[25] to represent the structure information of the image from the brightness, contrast and structure of the image, and it is more consistent with the sense of the human performance. Range of SSIM value is [0,1], and the higher the SSIM indicates the closer it is to the actual samples.

## 4.2    Work Platform Details

Detailed hardware and software are listed in Table 1. In order to ensure program running successfully, computing device must have enough memory at least 8Gb. Owing to huge time consumption in CPU, a method that utilizes CUDA to invoke GPU resources is adopted to speed up the training process.

**Table 1** Hardware and software platform.

| Name | Specification |
|---|---|
| CPU | Intel i7-6770K 4.0G Hz |
| RAM | DDR4 16GB |
| OS | Ubuntu 16.04 |
| GPU | Nvidia GTX 1080 |
| Framework | Pytorch 0.31 |

We use open source deep learning framework, pytorch 0.31, to build network and complete reconstruction.

## 4.3    Multi-scale and Single-scale for Training

A single model enabling to be implemented into multi-scale scenarios is critical for practical work. We consider model trained with upscaling factor=×2,×3,×4 (the corresponding training set/testset are $S_{train}, S_{test}$ ) have better PSNR and SSIM on different samples. The $S_{train}, S_{test}$ are constituted of samples which are randomly selected by same proportions from three scale datasets, respectively. In training stage, average MSE using with multi-scale, is higher than single-scale in the early epochs. Nevertheless, from Table 2, after convergence of both network, PSNR of multi-scale surpass counterpart 0.32dB and 0.46 dB in the case of $S_{test} = \{\times 2, \times 4\}$ and nearly be equal in $S_{test} = \{\times 3\}$ . We consider that using multi-scale training-set is more preferable. The following experiments are conducted by multi-scale training sets.

**Table 2** Comparison of PSNR using multi-scales and single-scale model for SR in different upscaling factors testsets.

| Method | Scale | | |
|---|---|---|---|
| | ×2 | ×3 | ×4 |
| 3DSRCNN(single) | 39.68 | 35.03 | 32.01 |
| 3DSRCNN(multi) | 40.00 | 35.02 | 32.47 |

### 4.4  Analysis of Training and Reconstruction

If training sets are blended up with multi-scale samples, the yielded model can be applied to different scale interpolation images which can save the cost of storing network. In Fig. 6, it is observed that the trend of the curves in different scale factors is basically same. There is slight fluctuation in some epoch, yet the overall curves are smooth and converges quickly. This is because the loss function of the network uses the MSE function, and MSE helps to train a higher network of PSNR values, but with SGD optimization, the loss function may fall into a bad local minimum, resulting in a slight drop in the PSNR during the training process. Note that not the low MSE value in training stage means that the reconstruction effect is better. In training phase, the calculation of MSE is based on one batch that is currently taken out from the training set, which can't be deemed as the criterion of reconstruction quality. It is supposed to use trained model to complete process of reconstruction to compare performance.

If more high frequency details were lost during downsampling, learning mapping relationship between low feature and high feature is relatively more difficult, thus brings intrinsic limitation on SR ability. It is evident to find there are higher gains in $S_{test} = \{\times 2\}$ and lower gains in the other testsets. Using the work platform of Table 1, it takes about 2.29 hours for each training epoch. In general, with the continuous iteration of the network, the PSNR will give higher value yet converge to a certain value due to the limitations of generalization ability of network and given training sets. In our work, it need about 20 epochs to converge.

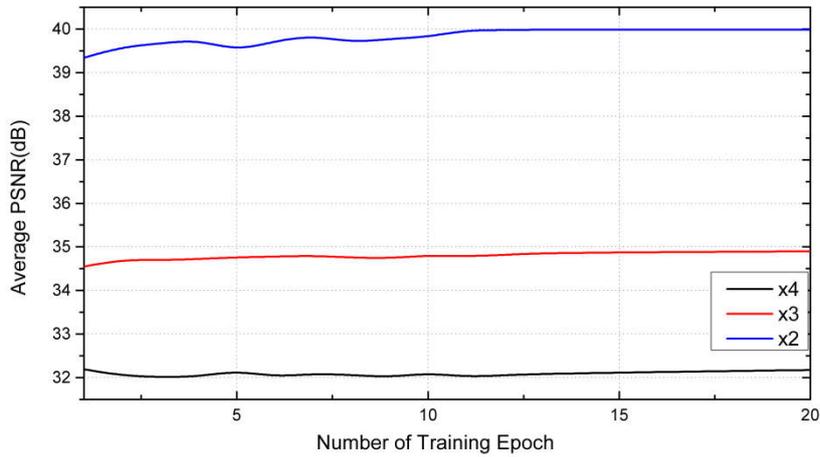

**Fig. 6** Average PSNR of 3DSRCNN on testsets with $\times 2$ $\times 3$ , $\times 4$ upscaling factors. The three curves are smooth, and we can find that residual learning can boost convergence.

The trained network model is actually a set of tensors storing the weight parameters of each neuron. Although trained CT image sets are divided into small blocks, in reconstruction stage, there is no requirement about the size of input images. The ultimate goal of our work is to acquire HR images by identical size, but if directly send the images with relatively large pixels to the network, it will consume a lot of memories (e.g. Reconstruction of data with size of $400 \times 400 \times 400$ takes up about 352G memory, tested in Pytorch 0.31). Therefore, we first crop input CT images to smaller sub-blocks with size of $100 \times 100 \times 100$ . Then, we assemble all reconnoitered sub-blocks to complete size HR images.

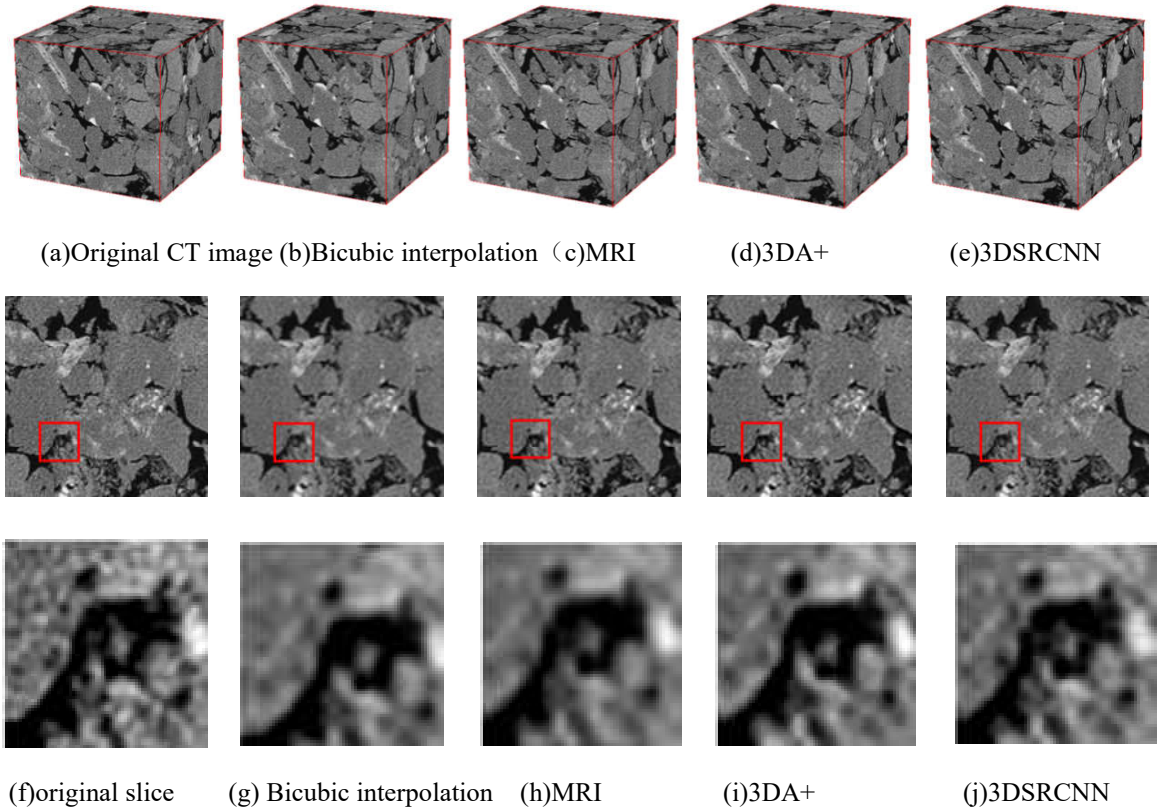

(a)Original CT image (b)Bicubic interpolation（c)MRI　(d)3DA+　(e)3DSRCNN

(f)original slice　(g) Bicubic interpolation　(h)MRI　(i)3DA+　(j)3DSRCNN

**Fig. 7** SR results of different algorithm in scale factor＝×4 testset of Sandstone rock sample. We can find that the zoom-in image of 3DSRCNN contains more clear texture than other method.

## 4.5　Training Parameters and Trade-offs

Network parameters, including network depth and convolution kernel size, will affect reconstruction accuracy and time. In most cases, increasing training epoch is conducive to a better performance. The mapping relationship that a network can learn from a given training set is yet limited to quantities of training data and structure of network. Appropriate increment of the network structure and training parameters is of crucial. In this section, we investigate the optimal setting to make a trade-off between performance and speed.

### 4.5.1　Depth of Network

For SR of single 2D images, Jiwon demonstrate the large depth help model to capture more contextual information and yield better performances than shallow ones. For training of 3D samples, the amount of computation and memory occupation is very large. Too many layers could slow down the convergence and exponentially improve the computational complexity. As shown in Fig. 8, we found that the PSNR will be improved until layers are added up to 12. After that, there is no apparent promotion on accuracy of reconstruction. Deepening the network is indeed effective, but we find that the is not the point, 'the deeper, the better', which needs to be explored through practical experiments based on given datasets. We try training with small number of datasets, so that the mapping relation learned from prior information is inherently limited, so an excessive increase of the layers will not substantially improve the accuracy of the reconstruction, Unexpectedly, it will bring about degradation problem due to over-fitting. If other parameters setting are uniform, we observe that training 20-layers for one epoch takes almost five times as much as 5-layers in Table 3.

**Table 3** Comparison of consuming time for training one epoch with different number of layers.

| Number of layers | 5 | 12 | 20 |
|---|---|---|---|
| Time(min) | 26.7 | 82.2 | 138.9 |

The 12-layers is about 0.15, 0.09 dB higher than 10-layers, 14-layers network in $S_{test} = \{\times 3, \times 4\}$, respectively. In addition, 20-layers is even much lower than that of the 12-layers in $S_{test} = \{\times 2\}$. On the other hand, PSNR of 12-layers is also relatively stable in whole testsets with different upscaling factors. These issues are also mentioned in Ref. 5, where improper increase of depth will cause degradation of SR accuracy, which also confirms our point that the number of layers of SR network should be proportional to the quantities of training data.

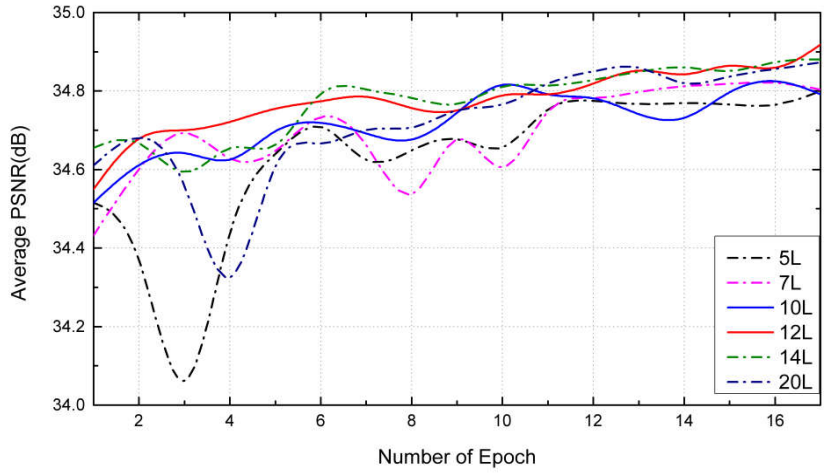

(a)  The average PSNR curves in  $S_{test} = \{\times 2\}$

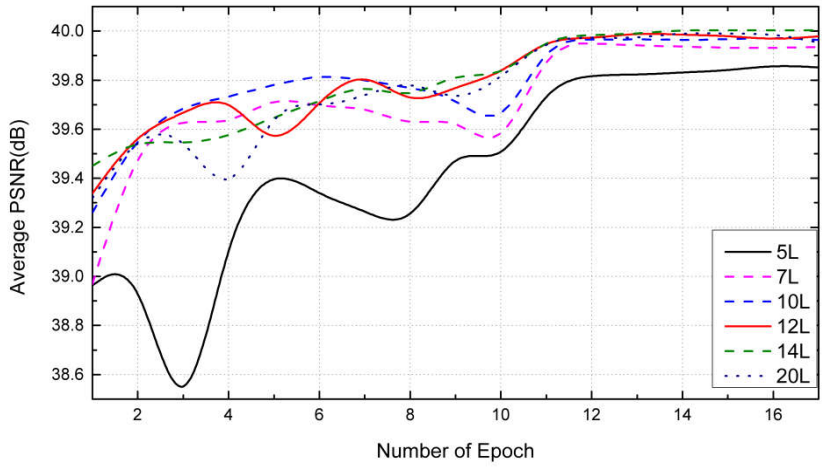

(b)  The average PSNR curves in  $S_{test} = \{\times 3\}$

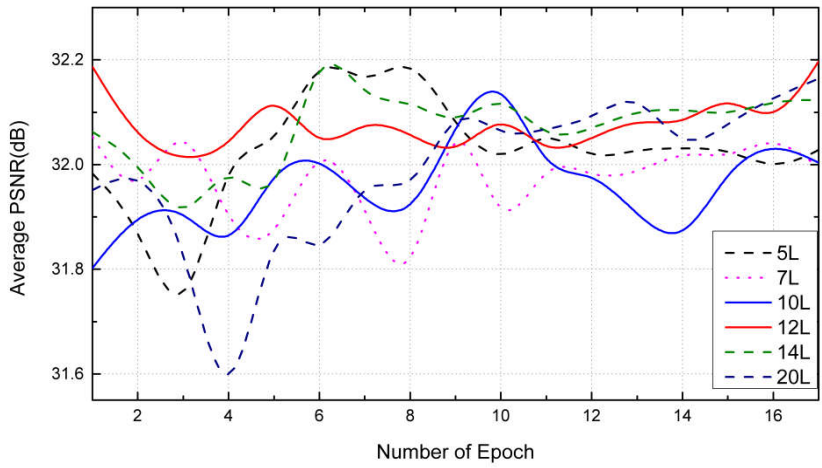

(c)  The average PSNR curves in  $S_{test} = \{\times 4\}$

### 4. 5. 2 Convolutional Kernel Size

In this subsection, we explore network sensitivity to different convolutional kernel size. In Ref. 13, all the filters of CNN are set as same size of $3 \times 3$. In Ref. 5, they examined the impact of different filter size, whereas their settings are inapplicable to ours due to the limitation of practical memory. Besides, the size of our input data is uniformly $25 \times 25 \times 25$, that means it is better to set relatively small filter. Based on previous experiment condition that number of layers is set to 12, we conduct three comparative trials with convolutional size of $3 \times 3 \times 3$, $5 \times 5 \times 5$ and $7 \times 7 \times 7$. The experimental results can be seen in Table 4. We find that there is no evident improvement when widening kernel size to $5 \times 5 \times 5$ than $3 \times 3 \times 3$. Instead, using $7 \times 7 \times 7$ has a large reduction in quality. By using small convolution kernel, the larger receptive field can be achieved by increasing the network depth as well. According to Formula (2), increasing the size of the filter incurs higher complexity than that of increasing the network depth. We consider that small filter size is more preferable.

**Table 4** Comparison of consuming time for training a epoch with different convolutional kernel size.

| Size(pixel) | $3 \times 3 \times 3$ | $5 \times 5 \times 5$ | $7 \times 7 \times 7$ |
|---|---|---|---|
| Time(min) | 81 | 250 | 428 |
| PSNR(dB) | 40.00 | 39.74 | 36.79 |

### 4. 5. 3 Residual vs Non-residual Learning

Increasing network layer could cause the problems of the gradient explosion/vanishing. We start an experiment: network layers are set to 12 and compare using residual learning with non-residual learning. It is observed that network with residual learning gives more outstanding performance shown in Fig. 6 and 9. In the early process of training, it is found that the MSE loss is very high. As shown in the Fig. 9, there is a dramatic wave during the training process of 5-10 epochs, even worse, PSNR has a large decline during the 7,8 epoch. Moreover, it will take long time to converge than using residual learning. Analogously, network with residual learning makes MSE converge to a smaller number in the first 3 epochs, and the network without residual learning needs more than 25 epochs to converge. The final result shows that using non-residual learning on $S_{\text{test}} = \{\times 3, \times 4\}$, is about 0.2dB, 0.1dB higher ,0.4dB lower than counterpart on $S_{test} = \{\times 2\}$. Using residual learning can enhance the stability of training process. When the network is deeper, the necessity of residual learning will be more prominent.

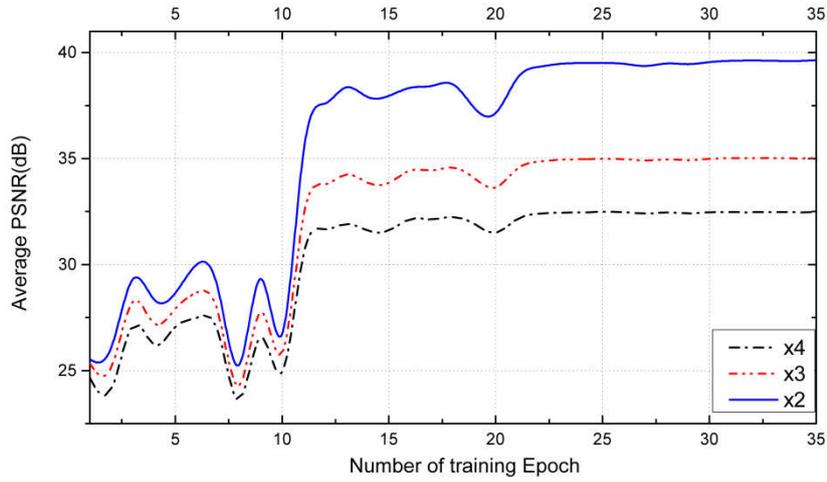

**Fig. 9** PSNR with non-residual learning. It's obvious to find that network with non-residual emerge shock waves in early stage. Non-residual learning will takes many epochs to maximum performance.

### 4.6    Comparison to State-of-the-art

The performance of our proposed network is experimentally analyzed compared with the benchmark bicubic interpolation, MRI[7], sparse-representation[11], 3D-A+[15], in the Table 4., the average PSNR increases 4.72dB, 2.48dB, 2.92dB, and SSIM increases 0.037, 0.05, 0.111 than input images (i.e. LR images), respectively. In addition, we can see that the 3DSRCNN training model have surpassed previous methods on three upscaling testsets with evaluation of PSNR. Compared 3DSRCNN with previous top performing algorithm 3DA+, ours is higher 1.42dB, 0.13dB, 0.82dB than it on $S_{test} = \{\times 2, \times 3, \times 4\}$. Therefore, it can be concluded that the neural network has a better learning ability than the traditional way. The appropriate setting of the network parameters can make full use of the redundant texture in the training image, thus learning more prior information and finally achieve excellent performance.

The reconstruction time and quality of 3DA+ is related to the selected size of the feature block. When the test feature block is large, the reconstruction quality is good. When the $3 \times 3 \times 3$ size block is selected, reconstructing the size of $400 \times 400 \times 400$ is about 22 minutes on CPU. Our work uses GPU to run program, and reconstruction of the same size image only need 3 minutes, besides ours have better performance of PSNR.

We selected regions with rich texture details to make visual contrast as enlarged areas. A selection of $70 \times 70$ pixels region, containing distinguishable texture, is zoomed-in to compare reconstruction quality of different methods, respectively. At the visual level, it can be seen 3DSRCNN contains relatively clear details (see Fig. 7), in contrast, the reconstructed images using other methods are more blurry.

**Table 5** The average results of PSNR(dB), SSIM in comparison with other algorithm. 3DSRCNN(single) is separately trained and only works for corresponding upscaling factor. In contrast, 3DSRCNN(multi) can be applied in different upscaling factors.

| Evaluation | Scale | Bicubic | MRI[7] | Sparse representation[11] | 3DA+[15] | 3DSRCNN(single) | 3DSRCNN(multi) |
|---|---|---|---|---|---|---|---|
| | 2 | 35.28 | 36.35 | 38.45 | 38.58 | 39.68 | **40.00** |
| PSNR(dB) | 3 | 32.54 | 33.15 | 34.74 | 34.89 | 35.03 | **35.02** |
| | 4 | 29.55 | 30.17 | 31.59 | 31.65 | 32.01 | **32.47** |
| | 2 | 0.950 | 0.965 | 0.983 | 0.983 | 0.986 | **0.987** |
| SSIM | 3 | 0.879 | 0.881 | 0.924 | 0.930 | 0.931 | **0.929** |
| | 4 | 0.738 | 0.799 | 0.817 | 0.828 | 0.850 | **0.849** |

## 5　Conclusion

In our work, we proposed a novel method, 3DSRCNN, based on deep learning to approach SR of voxel images. While using CNN to restore single LR image to high resolution have outperformed tradition method, there are many challenges previously mentioned to accomplish CT sample super-resolution. Our proposed model employ 3D-convolutional operation to handle CT images, which ensures the spatial continuity in plane and slice direction. Through practical experiments, We explored empirical guideline, such as learning rate, depth of network, as well as size of convolutional kernel, on designing network and parameters-tuning in training process. Moreover, stacking moderate network layers and adopting training strategies will exert on the accuracy and time of reconstruction. We found that training and reconstruction will occupy a lot of memory, which need to draw more attention. To cope with aforementioned issues, we crop original CT images to small blocks. We have demonstrated our method surpasses previous methods, which is shown in Table 5. For future research, we want to theoretically explain the effect of network depth on SR. Meanwhile, we will study the better training techniques to deal with the issues of larger quantities of 3D data, thus greatly reducing training time.

*Acknowledgements*

This work was supported by the National Natural Science Foundation of China (Grant No. 6137217 4).

**Yukai Wang** received his B.S. degree in electronics and information engineering from Dalian Maritime University, China, in 2017. He is currently working towards his master's degree at the College of Electronics and information Engineering, Sichuan University, Chengdu, China. His research interests include 3D reconstruction, image SR, deep learning.

**Qizhi Teng** is a professor and a senior member of the Chinese Institute of Electronics. She has published a book and a large number of papers, especially on image processing and image communication. She has undertaken nearly 30 completed and ongoing research projects sponsored by the National Natural Science Funding of China, PhD funding of the Education Department of China, Sichuan Provincial Natural Science Funding, and nationwide enter-prises. Her research interests include digital image processing, image communication, pattern recognition, and software engineering.

**Xiaohai He** received his B.S. and M.S. degrees in electrical engineering from Sichuan University, Chengdu, China, in 1985 and 1991, respectively. In 2002, he received his Ph.D. degree in biomedical engineering from the same university. He is currently a professor with the College of Electronics and Information Engineering, at Sichuan University. His research interests include image processing, pattern recognition, computer vision, image communication, and software engineering. He is a Senior Member of the Chinese Institute of Electronics. He is an editor of the Journal of Information and Electronic Engineering and an editor of the Journal of Data Acquisition & Processing.

**Junxi Feng** received his B.S. degree in electronics and information engineering from Sichuan University, China, in 2015. He is currently pursuing the Ph.D. degree in Communication and Information System from Sichuan University, Chengdu, China. His research interests include image processing, pattern recognition, machine learning, deep learning and their applications on petroleum geoscience.

**Tingrong Zhang** received her B.S. degree in electronics and information engineering from Sichuan University, China, in 2015. She is currently working towards his master's degree at the College of Electronics and information Engineering, Sichuan University, Chengdu, China. Her research interests include image restoration, medical images processing, deep learning and their applications on petroleum geoscience.